\documentclass[11pt]{article}
\usepackage{acl2004-colacl}
\usepackage{times}
\usepackage{amssymb}
\usepackage{epsfig}
\usepackage{xspace}
\usepackage{graphicx}

\newtheorem{definition}{Definition}

\newcommand{\set}[1]{\{#1\}}

\newcommand{\longversion}[1]{sentiment {#1}\xspace}
\newcommand{\Longversion}[1]{Sentiment {#1}\xspace}

\newcommand{\an}{analysis\xspace}

\newcommand{\An}{Analysis\xspace}
\newcommand{\Anl}{\Longversion{\an}}
\newcommand{\AnL}{\Longversion{\An}}
\newcommand{\detn}{detection\xspace}
\newcommand{\detr}{detector\xspace}
\newcommand{\detrs}{{\detr}s\xspace}
\newcommand{\Detn}{Detection\xspace}
\newcommand{\detnl}{subjectivity {\detn}\xspace}
\newcommand{\detrl}{subjectivity {\detr}\xspace}
\newcommand{\detrsl}{subjectivity {\detrs}\xspace}
\newcommand{\detnh}{subjectivity-{\detn}\xspace}

\newcommand{\Detrsl}{Subjectivity {\detrs}\xspace}
\newcommand{\DETNl}{Subjectivity {\Detn}\xspace}

\newcommand{\classification}{classification\xspace}
\newcommand{\classifier}{classifier\xspace}
\newcommand{\classifiers}{{\classifier}s\xspace}

\newcommand{\pol}{polarity\xspace}
\newcommand{\poll}{\longversion{\pol}}
\newcommand{\Pol}{Polarity\xspace}
\newcommand{\polclass}{{\pol} {\classification}\xspace}
\newcommand{\polclassr}{{\pol} {\classifier}\xspace}
\newcommand{\polclassrh}{{\pol}-{\classifier}\xspace}

\newcommand{\polclassrs}{{\pol} {\classifiers}\xspace}
\newcommand{\polclassh}{{\pol}-{\classification}\xspace}
\newcommand{\Polclass}{{\Pol} \classification\xspace}

\newcommand{\poldata}{{\pol} dataset\xspace}
\newcommand{\poldatah}{{\pol}-dataset\xspace}
\newcommand{\subjdata}{subjectivity dataset\xspace}
\newcommand{\Subjdata}{Subjectivity dataset\xspace}

\newcommand{\instance}{x}
\newcommand{\numinstances}{n}
\newcommand{\class}{C}
\newcommand{\classaindex}{1}\newcommand{\classbindex}{2}
\newcommand{\classa}{C_\classaindex}
\newcommand{\classb}{C_\classbindex}
\newcommand{\cost}{cost\xspace}

\newcommand{\Cost}{Cost\xspace}
\newcommand{\weightEng}{weight\xspace}
\newcommand{\weightsEng}{{\weightEng}s\xspace}

\newcommand{\preEstEng}{individual\xspace}
\newcommand{\PreEstEng}{Individual\xspace}
\newcommand{\preEst}{ind\xspace}
 
\newcommand{\assocEng}{association\xspace}
\newcommand{\AssocEng}{Association\xspace}
\newcommand{\assoc}{assoc\xspace}

\newcommand{\partcostEng}{partition cost\xspace}

\newcommand{\vertex}{v}
\newcommand{\graphitem}{\vertex}
\newcommand{\source}{s}
\newcommand{\sink}{t}

\newcommand{\nbsubjprob}{Pr^{NB}_{sub}}
\newcommand{\distthresh}{T}

\newcommand{\sigp}{p}

\newcommand{\procof}{}\newcommand{\procofthe}{}
\newcommand{\intl}{Intl.\xspace}
\newcommand{\conf}{Conf.\xspace}

\newcommand{\trans}{Trans.\xspace}
\newcommand{\transon}{\trans}
\newcommand{\workshop}{Wksp.\xspace}

\title{A Sentimental Education:  \AnL Using Subjectivity
  Summarization Based on Minimum Cuts}

\author{Bo Pang \and Lillian Lee \\ Department of Computer Science \\
  Cornell University \\ Ithaca, NY 14853-7501 \\ \{pabo,llee\}@cs.cornell.edu}

\date{}

\begin{document}
\maketitle
\begin{abstract}
{\em \Anl} seeks to identify the viewpoint(s) underlying a text span;
an example application is classifying a movie review as ``thumbs up''
or ``thumbs down''.  
To determine this
{\em \poll}, 
we propose a novel
machine-learning method that applies text-categorization techniques to
just the subjective portions of the document.
Extracting 
these portions can be implemented using efficient
techniques for finding {\em minimum cuts in graphs}; this greatly
facilitates incorporation of cross-sentence contextual constraints.

\medskip
\noindent {\bf Publication info:} {\em Proceedings of the ACL, 2004}.

\end{abstract}

\section{Introduction}
\label{sec:intro}

The computational treatment of 
opinion, sentiment, and subjectivity
has recently attracted a great deal of attention
\nocite{Hatzivassiloglou+McKeown:97a,Montes+Gelbukh+Lopez:99a,Das+Chen:01a,Subasic+Huettner:01a,Tong:01a,Dini+Mazzini:02a,Morinaga+al:02a,Pang+Lee+Vaithyanathan:02a,Turney:02a,Agrawal+al:03a,Dave+Lawrence+Pennock:03a,Durbin+Richter+Warner:03a,Liu+Lieberman+Selker:03a,Riloff+Wiebe+Wilson:03a,Yi+Nasukawa:03a,Yu+Hatzivassiloglou:03a,aaaisymp:04a}
(see references), in part
because of its potential applications.
For instance, information-extraction and
question-answering systems could flag statements 
and queries 
regarding
opinions
rather than facts \cite{Cardie+al:03a}.  Also,
it has proven useful for
companies, recommender systems, and 
editorial sites 
to 
create
summaries of people's experiences and opinions 
that consist of subjective expressions
extracted from reviews (as is commonly done in movie ads) or even just a review's {\em
polarity} --- positive (``thumbs up'') or negative (``thumbs down'').

Document \polclass poses a significant challenge to
data-driven methods,
resisting
traditional text-categorization techniques
\cite{Pang+Lee+Vaithyanathan:02a}.  Previous approaches  focused
on selecting indicative lexical features 
(e.g., 
the word ``good''),
classifying a document according to the number of such features that
occur anywhere within it.  In contrast, we propose the following
process: (1) label the sentences in the document as either
subjective or objective, discarding the latter; and then (2) apply a
standard machine-learning classifier to the resulting {\em extract}.
This can
prevent the \polclassr from considering irrelevant or even potentially
misleading text: for example,
although the sentence ``The protagonist tries to protect her good
name'' contains the word ``good'', it tells us nothing about the
author's opinion and in fact could well be embedded in a
negative movie review.
Also, as mentioned
above, 
subjectivity extracts can be provided to users as a  summary
of the 
sentiment-oriented content of the document.

Our results show 
that 
the subjectivity extracts 
we create
accurately
represent the sentiment information of the originating
documents 
in a much more compact form:  
depending on  choice of downstream \polclassr, 
we can achieve highly statistically significant improvement (from 
82.8\% to 86.4\%) or maintain the same level of performance
for the \polclass task
while 
retaining only 60\% of the reviews' words.
Also, we explore extraction methods
based on a {\em minimum cut} formulation,
which provides an efficient,  intuitive, and effective means for integrating
inter-sentence-level contextual information with traditional
bag-of-words features.

\section{Method}
\label{sec:method}

\subsection{Architecture} 

One can consider document-level \polclass to be just a special (more
difficult) case of
text categorization with sentiment- rather than topic-based 
categories.  Hence, standard machine-learning classification
techniques, such as support vector machines (SVMs),
can be applied to the entire documents themselves, as was done
by \newcite{Pang+Lee+Vaithyanathan:02a}. We refer to such
classification techniques as {\em default \polclassrs}.

\begin{figure}[ht]
\includegraphics[width=3.2in]{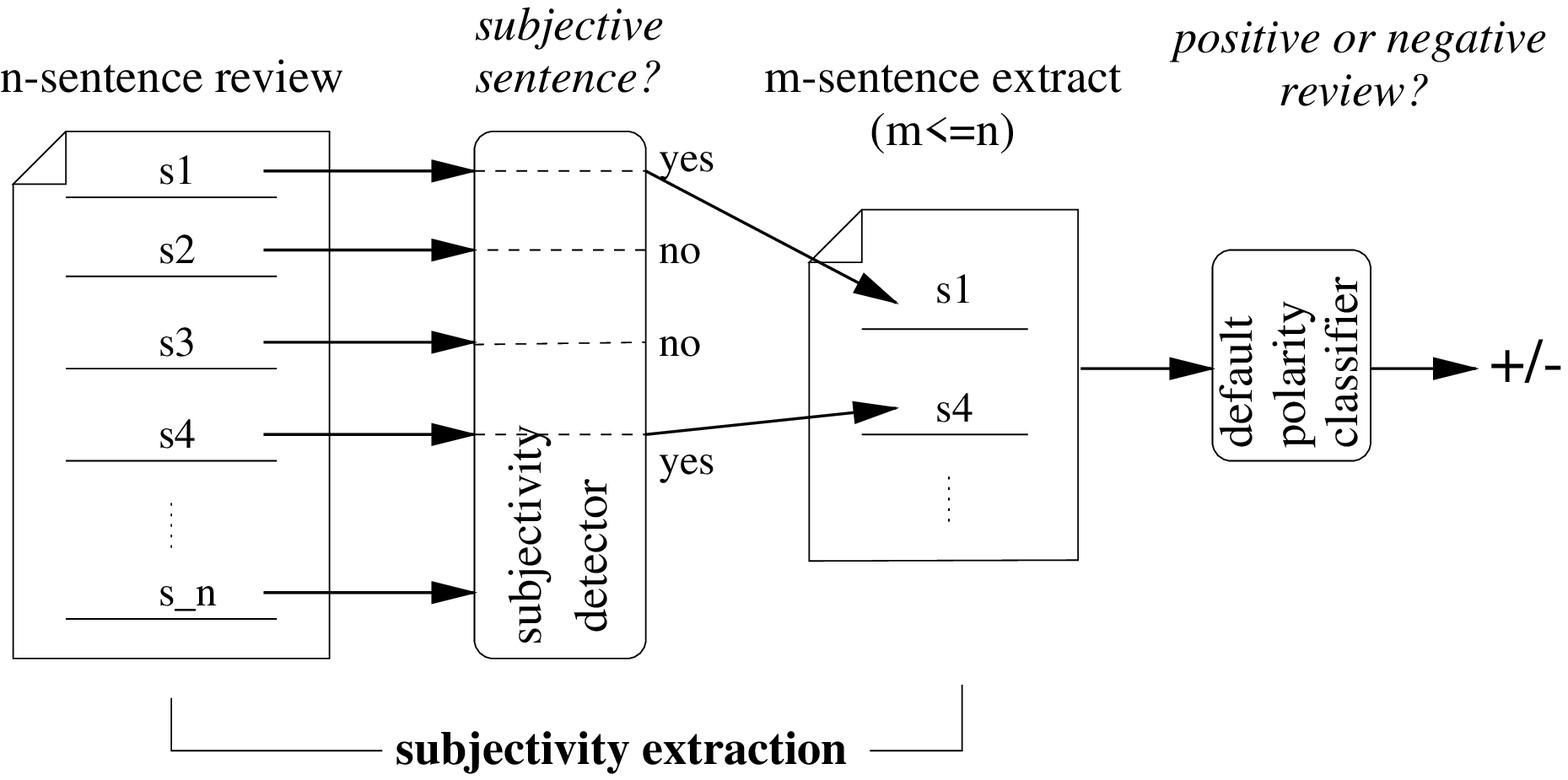}
\caption{\label{fig:struct}
{\small \Polclass via \detnl.}
}
\end{figure}

However, as noted above, we may be able to improve \polclass by 
removing objective sentences
(such as plot
summaries in a movie review).
We therefore propose, as depicted in Figure \ref{fig:struct},  to first employ a {\em {\detrl}} that
determines whether each sentence is subjective or not: 
discarding the objective ones
creates an {\em extract} that
should better represent a review's subjective content
to a default \polclassr. 

To our knowledge, previous work has not integrated sentence-level
\detnl with document-level \poll.  \newcite{Yu+Hatzivassiloglou:03a}
provide  methods for sentence-level analysis and for determining 
whether a document is subjective or not,
but do not combine these two
types of algorithms or consider document \polclass.
The motivation behind the single-sentence selection method of
\newcite{Beineke+al:04a} is to reveal a document's \poll, but they do
not evaluate the \pol-\classification accuracy that results.

\subsection{Context and \DETNl}
\label{sec:cuts}

As with document-level \pol \classification, we could perform
subjectivity detection on individual sentences by applying a standard
classification algorithm on each sentence in isolation.  However, 
modeling 
proximity relationships between sentences
would enable us to leverage {\em coherence}: text spans occurring near
each other (within discourse boundaries) may share the same
subjectivity status, other things being equal
\cite{Wiebe:94a}.

{ 
\newcommand{\y}{Y}
\newcommand{\m}{M}
\newcommand{\n}{N}
\newcommand{\inday}{.8}\newcommand{\indby}{.2}
\newcommand{\indam}{.5}\newcommand{\indbm}{.5}
\newcommand{\indan}{.1}\newcommand{\indbn}{.9}
\newcommand{\assocym}{1.0}
\newcommand{\assocyn}{.1}
\newcommand{\assocmn}{.2}
\begin{figure*}[t]
\begin{tabular}{lr}
\begin{minipage}{3.6in}
\includegraphics[width=3.3in]{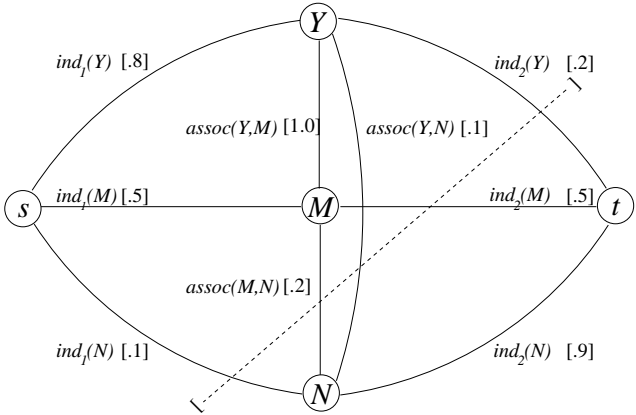}
\end{minipage}
&
\begin{minipage}{2.5in}
{\small
\begin{tabular}{l|lll}
$\classa$ & \PreEstEng & \AssocEng &
\Cost \\ 
 & penalties & penalties & \\ \hline
\set{\y,\m} & $\indby + \indbm + \indan$ & $\assocyn + \assocmn$ & 1.1 \\
(none) & $\inday + \indam + \indan$ & 0 & 1.4 \\
\set{\y,\m,\n} & $\indby + \indbm + \indbn$ & 0 & 1.6 \\
\set{\y} & $\indby + \indam + \indan$ & $\assocym + \assocyn$ & 1.9 \\
\set{\n} & $\inday + \indam + \indbn$ & $\assocyn + \assocmn$ & 2.5 \\
\set{\m} & $\inday + \indbm + \indan$ & $\assocym + \assocmn$ & 2.6 \\
\set{\y,\n} & $\indby + \indam + \indbn$ & $\assocym + \assocmn$ & 2.8 \\
\set{\m,\n} & $\inday + \indbm + \indbn$ & $\assocym + \assocyn$ & 3.3 
\end{tabular}
}
\end{minipage}
\end{tabular}
\caption{\label{fig:cutexample} {\small
Graph for classifying three
items.  Brackets enclose example values; here, the
\preEstEng scores happen
to be probabilities.  Based on {\em \preEstEng} scores
alone, we would put 
$\y$
(``yes'') in $\classa$,  $\n$ (``no'') in $\classb$, and be
undecided about $\m$ (``maybe'').  But the {\em \assocEng}
scores favor cuts that put $\y$ and $\m$ in the
same class, as shown in the table. Thus,
the minimum cut, indicated by the dashed line,
places $\m$ together with $\y$ in $\classa$.}
}
\end{figure*}

}

We would therefore like to supply our algorithms with {pair-wise}
interaction information,  e.g., to specify that  two particular sentences
should ideally receive the same subjectivity label {but not state
which label this should be}.
Incorporating such information is somewhat unnatural for classifiers
whose input consists simply of {\em individual} feature vectors, such
as Naive Bayes or SVMs, precisely because such classifiers label each
test item in isolation.  One could define synthetic features or
feature vectors to attempt to overcome this obstacle.  However, 
we propose an alternative that avoids the need for such feature
engineering:  we  use an efficient and intuitive  graph-based formulation relying on finding {\em
minimum cuts}.  
Our approach is inspired by \newcite{Blum+Chawla:01a}, although they focused on
similarity between items (the motivation being to combine labeled and
unlabeled data), whereas we are concerned with physical proximity between the
items to be classified; indeed, in computer vision, modeling proximity
information via graph cuts has led to very effective classification 
\cite{Boykov+Veksler+Zabih:99a}.

\subsection{Cut-based classification}
Figure \ref{fig:cutexample} shows a worked example of the
concepts in this section.

Suppose we have $\numinstances$ items $\instance_1, \ldots,
\instance_{\numinstances}$ to divide into two classes $\classa$ and
$\classb$, and we have access to two
types of information: 

 $\bullet$ {\em \PreEstEng} scores $\preEst_j(\instance_i)$:
non-negative estimates of each $\instance_i$'s preference for being in $\class_j$ based on just the features of
$\instance_i$ alone; and

 $\bullet$ {\em \AssocEng} scores $\assoc(\instance_i,
\instance_k)$:
non-negative estimates of how important it is that $\instance_i$ and
  $\instance_k$ be in  the same class.\footnote{Asymmetry is allowed, but we used symmetric
  scores.}

\smallskip

We would like
to maximize each item's ``net happiness'':  its \preEstEng score for the class it
is assigned to, minus its \preEstEng score for the other class.
But, we also want to penalize 
putting tightly-associated items into different classes.
Thus, after some algebra, we arrive at the following
optimization problem: assign the $\instance_i$s to
$\classa$ and $\classb$ so as to  minimize 
the {\em \partcostEng}
\newcommand{\tempomt}[1]{}
$$
\sum_{\instance\tempomt{_i} \in \classa}
\preEst_\classbindex(\instance\tempomt{_i}) +  \sum_{\instance\tempomt{_i} \in \classb}
\preEst_\classaindex(\instance\tempomt{_i})  + 
\sum_{\stackrel{\scriptstyle \instance_i \in \classa,}{\instance_k \in \classb}} \assoc(\instance_i,\instance_k).
$$

The problem appears
intractable,
since there are
$2^\numinstances$ possible binary partitions of the $\instance_i$'s.  
However, suppose we represent the
situation in the following manner.
Build an undirected graph $G$ with vertices
$\set{\graphitem_1, \ldots, \graphitem_\numinstances, \source,
\sink}$;  
the last two
are, respectively,  the {\em source} and {\em sink}.  Add
$\numinstances$ edges $(\source,\graphitem_i)$,
each with \weightEng $\preEst_1(\instance_i)$, and $\numinstances$
edges $(\graphitem_i, \sink)$, each with \weightEng
$\preEst_2(\instance_i)$.  Finally, add ${\numinstances \choose 2}$ edges
$(\graphitem_i,\graphitem_k)$, each  with \weightEng $\assoc(\instance_i,
\instance_k)$.  
Then, cuts in $G$ are defined as follows:
\begin{definition}
 A {\em cut} $(S,T)$ of $G$
is a partition of its nodes into 
sets $S = \set{\source}
\cup S'$ and $T = \set{\sink} \cup T'$, where $\source \not \in S',
\sink \not \in T'$.  Its {\em \cost} $\cost(S,T)$ 
 is the sum of the \weightsEng of all edges 
crossing from $S$ 
to $T$.  A {\em minimum cut} of $G$ is one of minimum
\cost.
\end{definition}
Observe that {every
cut 
corresponds to a partition 
of the items 
and
has \cost equal to the \partcostEng.
Thus,  
our
optimization problem reduces to finding minimum cuts}.

\medskip
\noindent {\bf Practical advantages} ~  As we have noted, 
formulating  our \detnh problem in
terms of graphs 
allows us to
model item-specific and pairwise information independently.  Note that
this is a very flexible paradigm.  For instance, it is perfectly
legitimate to use knowledge-rich
algorithms employing deep linguistic knowledge about sentiment
indicators to derive the \preEstEng scores.  And we could also
simultaneously use knowledge-lean methods to assign the \assocEng scores.
Interestingly, Yu and Hatzivassiloglou
\shortcite{Yu+Hatzivassiloglou:03a} compared an individual-preference
classifier against a relationship-based method, but didn't combine
the two; {the ability to {\em coordinate} such algorithms is {
precisely} one of the strengths of our approach.}

But a crucial advantage specific to the utilization of a
minimum-cut-based approach
is that 
we can use {\em maximum-flow} algorithms with {polynomial}
asymptotic running times --- and {near-linear} running times in
practice --- 
to {\em exactly} compute the minimum-cost cut(s), despite the apparent
intractability of the optimization problem
\cite{CLR,Ahuja+Magnanti+Orlin:93a}.\footnote{Code available at 
http://www.avglab.com/andrew/soft.html.}  In contrast, other
graph-partitioning problems that have been previously used to
formulate NLP classification problems\footnote{Graph-based
approaches to general {\em clustering} problems are too numerous to
mention here.}  are NP-complete
\cite{Hatzivassiloglou+McKeown:97a,Agrawal+al:03a,Joachims:03a}.

\section{Evaluation Framework}
\label{sec:filterproblem}

Our experiments involve classifying movie reviews as either
positive or negative, an appealing task for several reasons.
First, as mentioned in the introduction, providing \pol information
about reviews is a useful service: witness the popularity of
www.rottentomatoes.com.  Second, 
movie
reviews 
 are apparently harder to classify than reviews of other products
\cite{Turney:02a,Dave+Lawrence+Pennock:03a}. 
Third, the correct label can be extracted automatically from rating
information
(e.g., number of stars).
Our data\footnote{Available at
  www.cs.cornell.edu/people/pabo/movie-review-data/ (review corpus version 2.0).} contains 1000 positive and 1000 negative reviews 
all written before 2002,  with a cap of 20 reviews per author 
(312 authors total) 
per category.
We refer to this corpus
as the {\bf \poldata}.

\medskip
\noindent {\bf Default \polclassrs} ~ 
We tested support vector machines (SVMs) and Naive Bayes (NB).  
Following Pang et al. \shortcite{Pang+Lee+Vaithyanathan:02a}, we use
unigram-presence features: the $i$th coordinate
of a feature vector is 1 if the corresponding unigram occurs in the
input text, 0 otherwise.  (For SVMs, the feature vectors are length-normalized).
Each default document-level \polclassr is
trained and tested on the extracts formed by applying one of the
sentence-level \detrsl to 
reviews in the \poldata.  

\medskip
\noindent {\bf \Subjdata} ~
To train our \detrs, we need a collection of
labeled sentences. 
Riloff and Wiebe \shortcite{Riloff+Wiebe:03a} state that
``It is [very hard]
to obtain collections of individual sentences that
can be easily identified as subjective or objective''; 
the \poldatah sentences,  for example, have not been so
annotated.\footnote{We therefore could not directly evaluate 
sentence-\classification accuracy  on the \poldata.}
Fortunately, we were able to mine the Web to create a 
large, automatically-labeled sentence corpus\footnote{Available at
  www.cs.cornell.edu/people/pabo/movie-review-data/ , sentence corpus
  version 1.0.}. 
To gather subjective sentences (or phrases), 
we collected 5000 
movie-review {\em snippets} (e.g., ``bold,
imaginative, and impossible to resist'') 
from
www.rottentomatoes.com.
To obtain (mostly) objective data, we took 5000 sentences from plot summaries
available from the Internet Movie Database (www.imdb.com).
We only selected sentences or snippets at
least ten words long and drawn from reviews or plot summaries of movies released
post-2001, which prevents overlap with the \poldata.

\begin{figure*}[ht]
\includegraphics[width=6.2in]{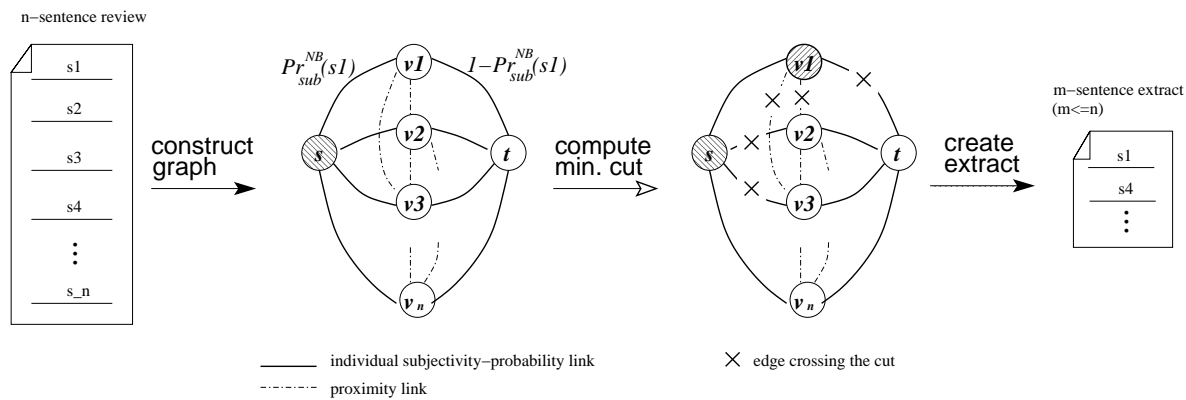}
\caption{\label{fig:filter} 
Graph-cut-based creation of subjective extracts.
}
\end{figure*}

\medskip
\noindent {\bf \Detrsl} ~
As noted above, we can use our default \polclassrs  as 
``basic'' sentence-level \detrsl (after 
retraining on the \subjdata)
to produce extracts of the  original reviews.
We also create a family of cut-based \detrsl; these
take as input the
{\em set} of sentences appearing in a single document and
determine the subjectivity status of all the sentences
simultaneously using per-item  and pairwise relationship
information. 
Specifically, for a given 
document, we use the construction in Section
\ref{sec:cuts} to build a graph 
wherein the source $s$ and sink $t$ 
correspond to the class of subjective and objective sentences,
respectively, and each internal node $\vertex_i$ 
corresponds to the document's $i^{th}$
sentence $s_i$.
We can set the \preEstEng scores $\preEst_1(s_i)$ to 
$\nbsubjprob(s_i)$ and
$\preEst_2(s_i)$  to $1 - \nbsubjprob(s_i)$, as shown in  Figure
\ref{fig:filter}, where $\nbsubjprob(s)$  denotes Naive Bayes' estimate  of the probability
that sentence $s$ is subjective; or, we can use the weights produced by
the SVM classifier instead.\footnote{We converted SVM output $d_i$, which is a signed
  distance (negative=objective)
  from the separating hyperplane, to non-negative numbers by
$$
\preEst_1(s_i) \stackrel{def}{=} \cases{ 
1 & $d_i > 2$; \cr
(2 + d_i)/4 & $-2 \le d_i \le 2$; \cr
0 & $d_i < -2$.}
$$
and $\preEst_2(s_i) = 1 - \preEst_1(s_i)$.
Note that scaling is employed only for consistency; the algorithm itself does not require probabilities for \preEstEng scores.
}
If we  set all the  \assocEng scores to zero, then the minimum-cut classification of the
sentences is the same as that of the basic subjectivity detector.
Alternatively, we
incorporate the degree
of {\em proximity} between pairs of sentences, controlled by three
parameters.  The threshold $\distthresh$ specifies the maximum
distance two sentences can be separated by and still be considered
proximal.
The non-increasing function $f(d)$ specifies how the influence of proximal sentences decays with
respect to distance $d$; in our experiments, we tried $f(d)=1$,
$e^{1-d}$, and
$1/d^2$.
The constant $c$ controls the relative influence of the \assocEng
scores:
a larger $c$
makes the minimum-cut algorithm  more loath to put proximal sentences in
different classes.
With these in hand\footnote{Parameter training is driven by optimizing the
performance of the downstream \polclassr rather than the \detr
itself  because the \subjdata's sentences come from different
reviews, and so are never proximal.}, we set (for $j > i$)
$$
\assoc(s_i,s_j) \stackrel{def}{=} \cases{ f(j-i) \cdot c & if $(j-i) \le
\distthresh$; \cr
	0 & otherwise.}
$$

\newcommand{\wholeReview}{{\it Full review}\xspace}
\newcommand{\Extract}[1]{{\it $\mbox{Extract}_{\mbox{\small{#1}}}$}\xspace}
\newcommand{\NBExtract}{\Extract{NB}} 
\newcommand{\NBgraphExtract}{\Extract{NB+Prox}}
\newcommand{\SVMExtract}{\Extract{SVM}}
\newcommand{\SVMgraphExtract}{\Extract{SVM+Prox}}
\newcommand{\flippedNBExtract}{${\mathbf\neg}$\NBExtract}
\newcommand{\better}[1]{{\it #1}}
\newcommand{\worse}[1]{#1}

\section{Experimental Results}
\label{sec:eval}

Below, we  report average accuracies computed by ten-fold
cross-validation over the \poldata.  
Section \ref{sec:eval:basic} examines our basic subjectivity extraction algorithms, which are based on individual-sentence predictions alone.
Section \ref{sec:eval:context} evaluates the more sophisticated
form of subjectivity extraction that
incorporates context information via the minimum-cut paradigm.

As we will see, the use of subjectivity extracts can in the best case
provide satisfying improvement in \polclass, and otherwise
can at least 
yield \polclassh accuracies indistinguishable
from employing the full review.  At
the same time, the extracts we create are both smaller on average
than the original document and more effective as input to a default
\polclassr than the same-length counterparts produced by standard
summarization tactics (e.g., first- or last-N sentences).  We
therefore conclude that subjectivity extraction 
produces effective summaries of document sentiment.

\subsection{Basic subjectivity extraction}
\label{sec:eval:basic}
As noted in Section \ref{sec:filterproblem}, both Naive Bayes and SVMs
can be trained on our \subjdata and then used as a basic \detrl.
The former
has somewhat better
average ten-fold cross-validation performance on the \subjdata (92\%
vs. 90\%), and so for space reasons, our initial discussions will
focus on the results attained via NB \detnl.

\begin{figure*}[ht]
{ 
\newcommand{\adjplotwidth}{3.5in}
\hspace*{-.4in}\begin{tabular}{cc}
\includegraphics[width=\adjplotwidth]{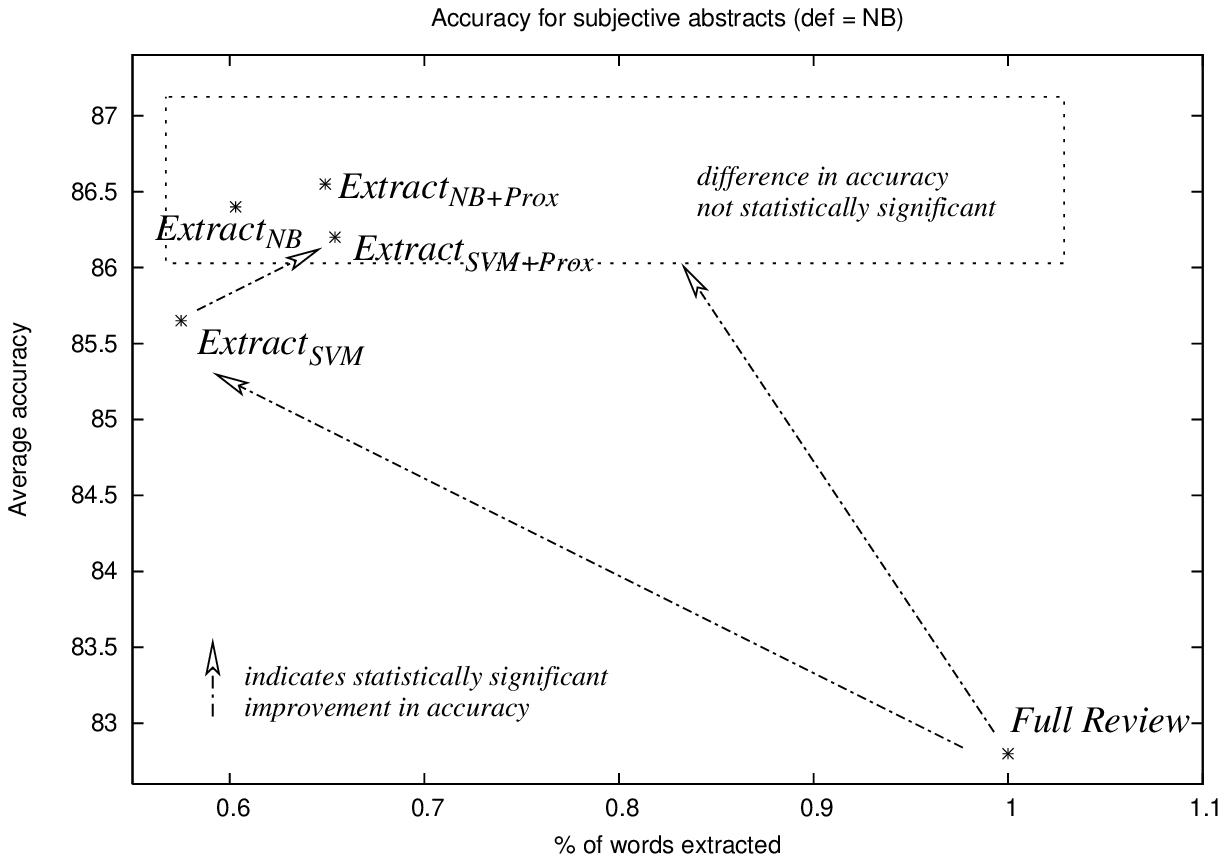}
\includegraphics[width=\adjplotwidth]{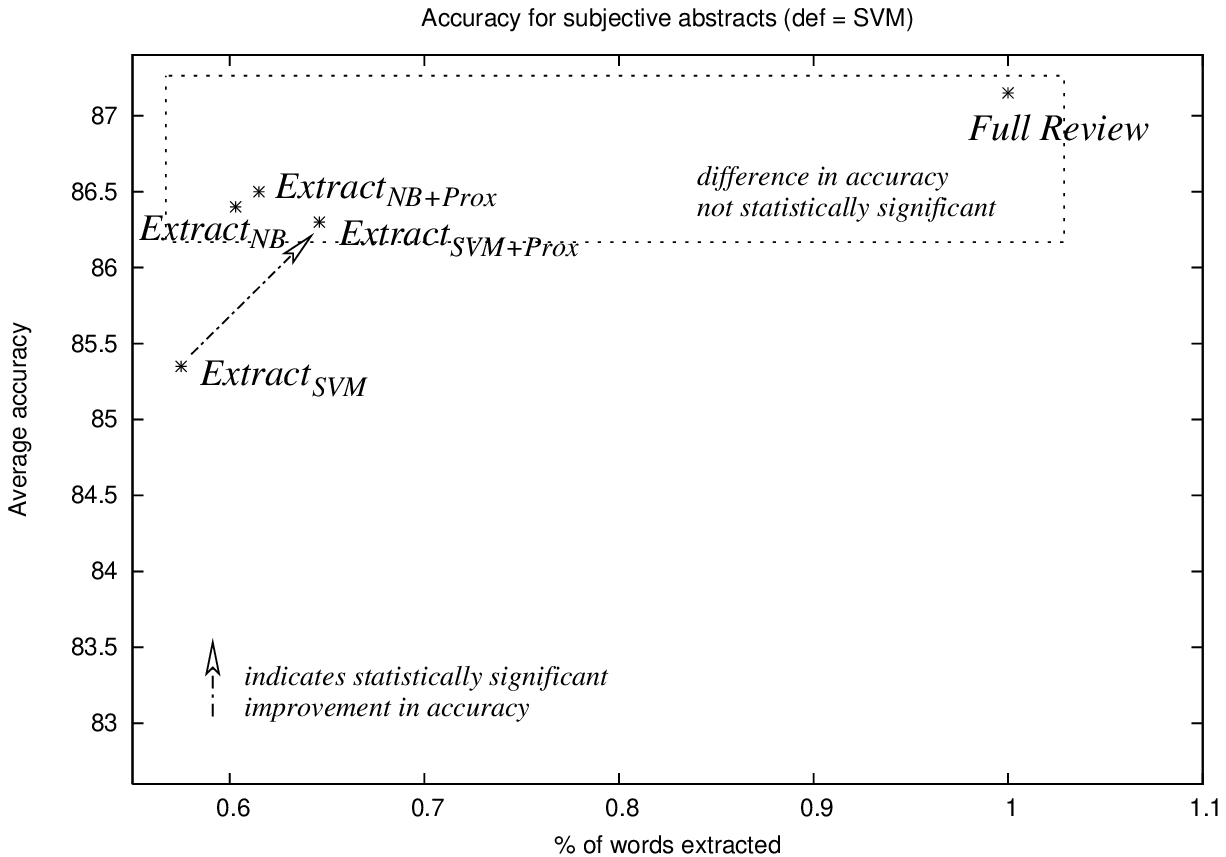}
\end{tabular}
\caption{\label{fig:preserve-vs-acc}
Word
preservation rate vs. accuracy, NB (left) and SVMs (right)
as default \polclassrs.  Also indicated are results for some
statistical significance tests.}
} 
\end{figure*}

Employing Naive Bayes as a \detrl (\NBExtract) in conjunction with a
Naive Bayes document-level \polclassr achieves 86.4\%
accuracy.\footnote{This result and others are depicted in Figure
\ref{fig:preserve-vs-acc}; for now, consider only the y-axis in those plots.}
This is a clear improvement over the 82.8\% that results when no
extraction is applied (\wholeReview); indeed, the difference is highly
statistically significant ($\sigp < 0.01$, paired t-test).  With SVMs
as the
\polclassr instead, the \wholeReview performance rises 
to 87.15\%, but  comparison via the paired t-test
reveals that this  is statistically indistinguishable from
the 86.4\% that is achieved by running the SVM \polclassr on 
\NBExtract input. (More improvements to extraction performance are
reported later in this section.)

\begin{figure*}[ht]
\newcommand{\adjplotwidth}{3.5in}
\hspace*{-.4in}\begin{tabular}{cc}
\includegraphics[width=\adjplotwidth]{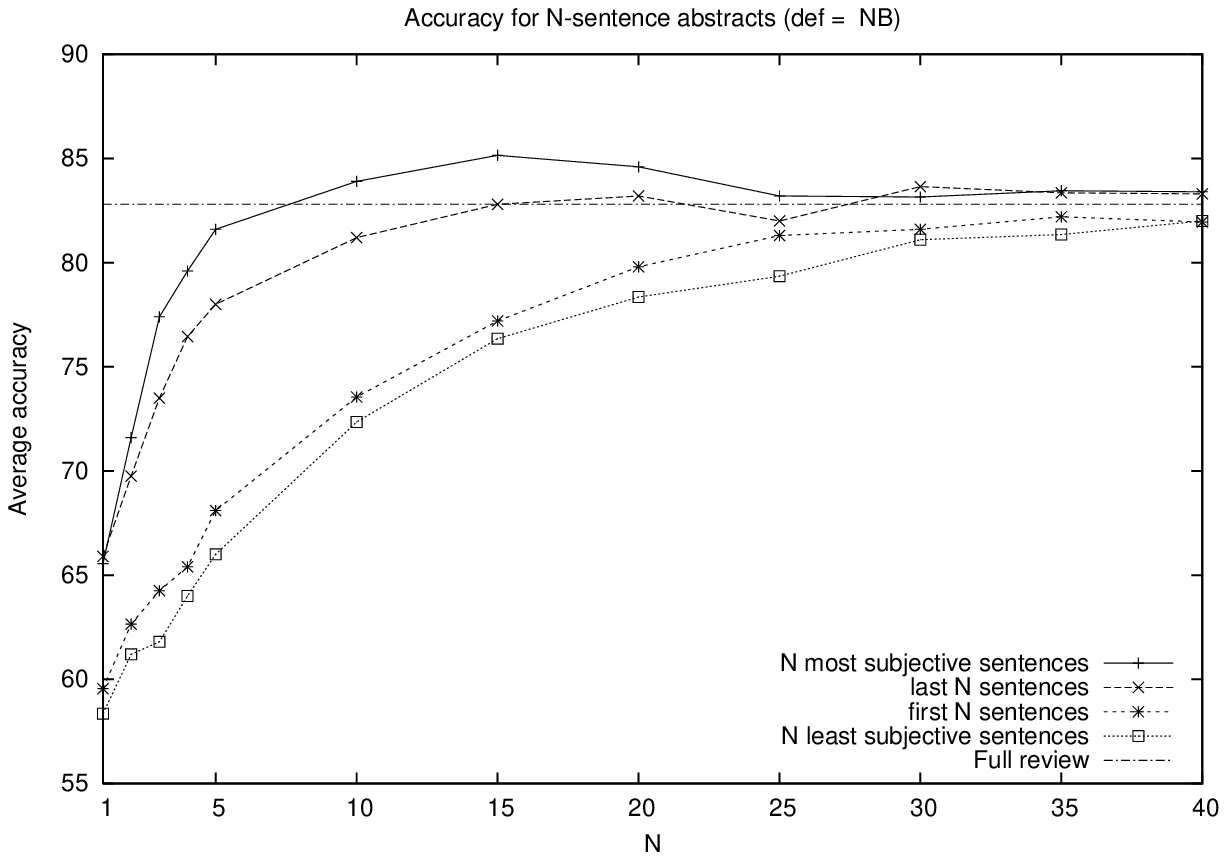}
\includegraphics[width=\adjplotwidth]{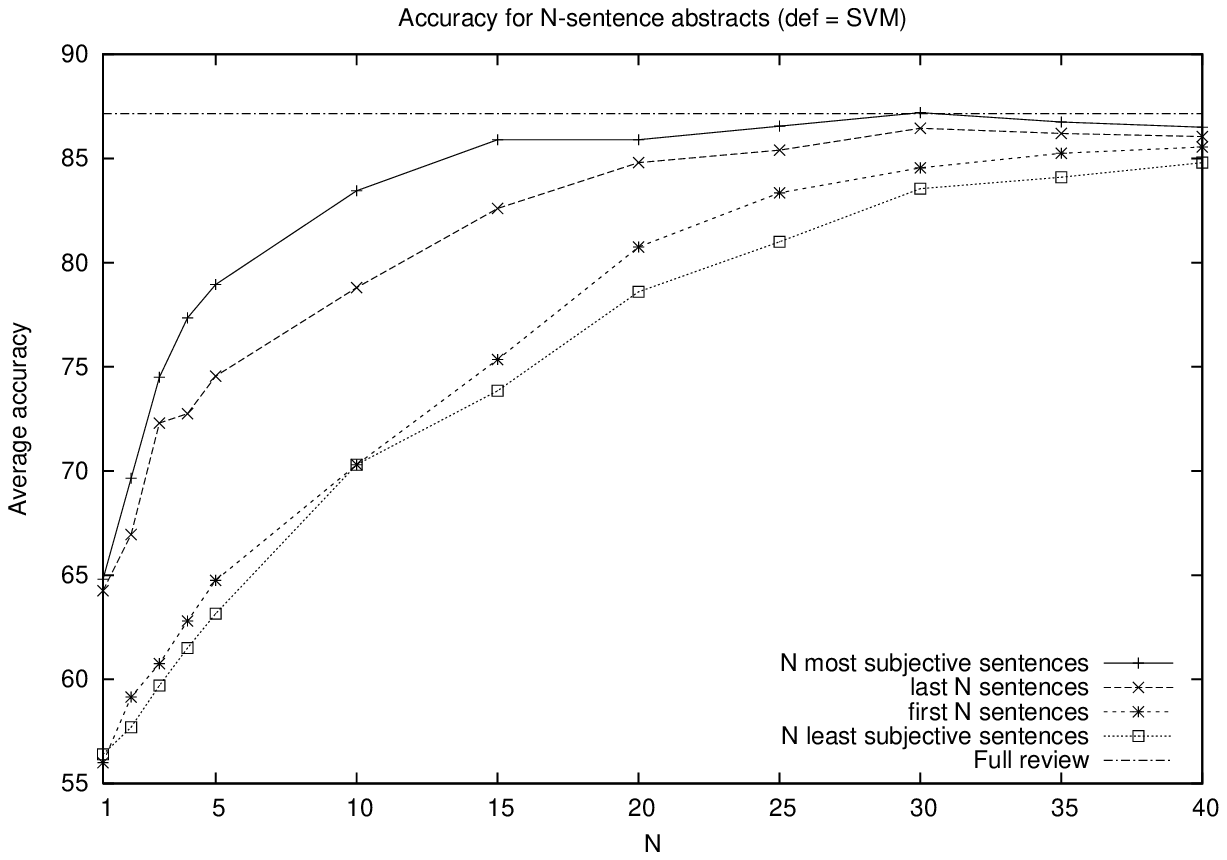}
\end{tabular}
\caption{\label{fig:n-sent} 
Accuracies using N-sentence extracts for NB (left) and SVM (right)
default \polclassrs.  
}
\end{figure*}

These findings indicate\footnote{Recall that direct evidence is not
 available because the \poldata's sentences lack subjectivity labels.} that the extracts  preserve (and, in the
NB \polclassrh case, apparently clarify) the sentiment information in the
 originating documents, and thus are good summaries from the
 \polclassh point of view.
Further support comes from a ``flipping'' experiment: if we give as
input to the default \polclassr an extract consisting of the sentences
labeled  {\em objective},
accuracy drops dramatically to 71\% for NB and
67\% for SVMs.  This confirms our hypothesis that sentences discarded
by the subjectivity extraction process are indeed much less indicative
of \poll.

Moreover, 
the subjectivity extracts are much more compact than the original documents (an
important feature for a summary to have): they contain on
average only about 60\% of the source reviews' words.  (This
{\em word preservation rate} is
plotted along the x-axis in the graphs in Figure \ref{fig:preserve-vs-acc}.)
This prompts us to study how much reduction of the original documents
\detrsl can perform and still accurately represent the texts'
sentiment information.

We can create subjectivity extracts of varying lengths by taking just
the $N$ {\em most subjective} sentences\footnote{These are the $N$ sentences assigned the highest probability by the
basic NB \detr, regardless of whether their probabilities exceed
50\% and so would actually  be classified as subjective by Naive
Bayes.  
For reviews with fewer than $N$ sentences, the entire review will be 
returned.} from the originating review.  
As one baseline to compare against, we
take the canonical summarization standard of extracting the  {\em
first $N$} sentences --- in general settings, authors often begin
documents with an overview.  We also consider the  {\em last $N$}
sentences: in many documents, concluding material may be a good summary, and
www.rottentomatoes.com tends to select ``snippets'' 
from the end of movie reviews \cite{Beineke+al:04a}.  Finally, as a
sanity check, we include results from the $N$ {\em least subjective}
sentences according to Naive Bayes.

Figure \ref{fig:n-sent} shows the \polclassr results as $N$ ranges
between 1 and 40.  Our first observation is that the NB \detr
provides very good ``bang for the buck'': with subjectivity extracts
containing as few as 15 sentences,
accuracy is quite close to what one gets if the entire review is used.
In fact, for the NB \polclassr, just using the 5 most subjective sentences is
almost as informative as the \wholeReview  
while containing on average only about 22\% of the source reviews' words.

Also, it so happens that at $N=30$, performance is actually slightly
better than (but statistically indistinguishable from)
\wholeReview even when the SVM default \polclassr is used (87.2\%
vs. 87.15\%).\footnote{Note that roughly half of the documents in the \poldata contain more than 30
sentences (average=32.3, standard deviation 15).}
This suggests potentially effective extraction alternatives 
other than using a fixed probability threshold (which resulted in the
lower accuracy of 86.4\% reported above).

Furthermore, we see in Figure \ref{fig:n-sent} that the $N$
most-subjective-sentences method generally outperforms the other
baseline summarization methods (which perhaps suggests that sentiment
summarization cannot be treated the same as topic-based summarization,
although this conjecture would need to be verified on other domains
and data).  
It's also interesting to observe how much better the last $N$ sentences
are than the  first $N$ sentences;
this may reflect a (hardly surprising) tendency for movie-review authors to 
place
plot descriptions at the beginning rather than the end of the text
and 
conclude with overtly opinionated statements.

\subsection{Incorporating context information}
\label{sec:eval:context}
The previous section demonstrated the value of \detnl. We now examine
whether context information, particularly regarding sentence
proximity, can further improve subjectivity extraction. As discussed
in Section \ref{sec:cuts} and \ref{sec:filterproblem}, contextual constraints are easily
incorporated via the minimum-cut formalism but are 
not natural inputs
for standard Naive Bayes and SVMs.

Figure \ref{fig:preserve-vs-acc} shows the effect of 
adding in proximity information.  \NBgraphExtract and \SVMgraphExtract
are the graph-based
\detrsl using Naive Bayes and SVMs, respectively, for the \preEstEng scores;
we depict the best performance achieved by a single
setting of the three proximity-related edge-weight parameters over all ten data
folds\footnote{Parameters are chosen from $\distthresh \in \{1,2,3\}$, $f(d)
\in \{1, e^{1-d}, 1/d^2\}$, and $c \in [0,1]$ at
intervals of 0.1.} (parameter selection was not a focus of the current
work).  The two comparisons we are most
interested in are \NBgraphExtract versus \NBExtract and
\SVMgraphExtract versus \SVMExtract.

We see that the context-aware graph-based
\detrsl tend to create extracts that are more informative
(statistically significant so (paired
t-test) for SVM
\detrsl only), although these extracts are longer than their context-blind
counterparts.  We note that the performance enhancements cannot be attributed
entirely to the mere inclusion of more sentences regardless of whether
they are subjective or not ---  one counterargument is that 
\wholeReview  yielded
substantially worse results for the NB default \polclassr ---   and at any rate, the
graph-derived extracts are still
substantially more concise than the full texts.

Now, while incorporating a  bias for assigning nearby sentences to the same
category  into NB and SVM
\detrsl seems to require some non-obvious feature engineering, we also wish to investigate whether our graph-based paradigm
makes better use of  contextual constraints that {\em can} be (more or
less) easily encoded
into the input of standard classifiers.
For illustrative purposes, we consider paragraph-boundary information,
looking only at  SVM \detnl for simplicity's sake.

It seems intuitively plausible that 
paragraph boundaries (an approximation to discourse boundaries) loosen
coherence constraints between nearby sentences.  To capture this
notion for minimum-cut-based classification, we can simply reduce the
\assocEng scores for all pairs of sentences that occur in different
paragraphs by multiplying them by a {cross-paragraph-boundary weight}
$w \in [0,1]$.  For standard \classifiers,  
we can employ the trick of having the \detr treat paragraphs, rather
than sentences, as the basic unit to be labeled.  This enables the
standard \classifier to utilize coherence between sentences in the
same paragraph; on the other hand, it also (probably unavoidably)
poses a hard constraint that all of a paragraph's sentences get the
same label, which increases noise sensitivity.\footnote{For example,
in the
data we used, 
boundaries may have been missed due to malformed html.}  Our experiments reveal the graph-cut
formulation 
to be the better approach: 
for both default \polclassrs (NB and SVM),
some choice of parameters (including $w$) for \SVMgraphExtract yields statistically
significant improvement over its paragraph-unit non-graph counterpart
(NB: 86.4\% vs. 85.2\%; SVM: 86.15\% vs. 85.45\%).

\section{Conclusions}

We examined the 
relation between \detnl and \polclass, showing
that \detnl can compress reviews into much shorter extracts 
that still retain polarity information at a level comparable to that of the full review.
In fact, for the Naive Bayes \polclassr, the subjectivity extracts are
shown to be 
more effective input
than the originating document, which suggests that they are not only
shorter, but also ``cleaner'' representations of the intended
polarity.

We have also shown that employing the minimum-cut framework results in the
development of efficient 
algorithms for sentiment
analysis.  
Utilizing contextual information via this framework
can lead to statistically significant
improvement in \polclassh accuracy.  
Directions for future research include developing parameter-selection
techniques, incorporating other sources of contextual cues besides
sentence proximity, and investigating other means for modeling such
information.

\section*{Acknowledgments}

We thank Eric Breck, Claire Cardie, Rich Caruana, Yejin Choi, Shimon
Edelman, Thorsten Joachims, Jon Kleinberg, Oren Kurland, Art Munson,
Vincent Ng, Fernando Pereira, Ves Stoyanov, Ramin Zabih, and the
anonymous reviewers for helpful comments.  This paper is based upon
work supported in part by the National Science Foundation under grants
ITR/IM IIS-0081334 and IIS-0329064, a Cornell Graduate Fellowship in Cognitive Studies, and by an Alfred P. Sloan Research
Fellowship. Any opinions, findings, and conclusions or recommendations
expressed above are those of the authors and do not necessarily
reflect the views of the National Science Foundation or Sloan
Foundation.

\bibliographystyle{fullname}
{

}

\end{document}